\documentclass[runningheads]{llncs}
\usepackage[T1]{fontenc}
\usepackage{amsmath}
\usepackage{multirow}
\usepackage{hyperref}
\usepackage{subcaption}
\usepackage{graphicx,array, rotating, ltablex, xspace, amssymb, mathtools}
\usepackage{url}
\newcommand{\keywords}[1]{\par\addvspace\baselineskip
\noindent\keywordname\enspace\ignorespaces#1}

\begin{document}

\mainmatter  

\title{An Improved Subject-Independent Stress Detection Model Applied to Consumer-grade Wearable Devices}

\titlerunning{Improved Subject-independent Stress Detection Model}

%
%

\institute{Dublin City University, Ireland \and
VNU-HCM, University of Science, Vietnam
}

\author{Van-Tu Ninh\inst{1,}\thanks{Both authors contributed equally to this research.}
\and
Manh-Duy Nguyen\inst{1, \star} \and
Sinéad Smyth\inst{1} \and
Minh-Triet Tran\inst{2} \and
Graham Healy\inst{1} \and
Binh T. Nguyen\inst{2} \and
Cathal Gurrin\inst{1}}
%
\authorrunning{Van-Tu Ninh, Manh-Duy Nguyen, et al.}

%

%



%
%

\maketitle

\begin{abstract}
Stress is a complex issue with wide ranging physical and psychological impacts on human daily performance.
Specifically, acute stress detection is becoming a valuable application in contextual human understanding.
Two common approaches to training a stress detection model are subject-dependent and subject-independent training methods. 
Although subject-dependent training methods have proven to be the most accurate approach to build stress detection models, subject-independent models are a more practical and cost-efficient method, as they allow for the deployment of stress level detection and management systems in consumer-grade wearable devices without requiring training data for the end user.
To improve the performance of subject-independent stress detection models, in this paper, we introduce a stress-related bio-signal processing pipeline with a simple neural network architecture using statistical features extracted from multimodal contextual sensing sources including Electrodermal Activity (EDA), Blood Volume Pulse (BVP), and Skin Temperature (ST) captured from a consumer-grade wearable device.
Using our proposed model architecture, we compare the accuracy between stress detection models that  use measures from each individual signal source, and one model employing the fusion of multiple sensor sources.
Extensive experiments on the publicly available WESAD dataset demonstrate that our proposed model outperforms conventional methods as well as providing $1.63\%$ higher mean accuracy score compared to the state-of-the-art model while maintaining a low standard deviation. 
Our experiments also show that combining features from multiple sources produces more accurate predictions than using only one sensor source individually.

\end{abstract}

\section{Introduction}
The development of sensor technology in recent years has lead to the availability of both consumer-grade and medical-grade wearable devices which has facilitated research into personal sensing with applications in self-quantification, lifelogging, and healthcare \cite{Gurrin2014LifeloggingPB}.
This has also resulted in the creation of many large multimodal personal datasets \cite{Gurrin2014LifeloggingPB} comprising of different data types (e.g, passive visual capture, mobile device context, physiological data) \cite{Gurrin2014APB} that enables research community to develop intelligent systems to track individual's health and gain more insights of an individual's personal data such as daily-life event segmentation \cite{Gurrin2016OverviewON}, activities of daily-living identification as an indicator in health tracking systems \cite{Gurrin2019OverviewOT}, etc.
Although multiple data sources are recorded in multimodal personal datasets \cite{Gurrin2021}, only the combination of visual and related metadata including semantic locations, daily-life activities, date and time are employed extensively in research \cite{Gurrin2019OverviewOT,Ninh2020OverviewOI} while others has not yet been exploited.
Typically, physiological signals are usually ignored due to the limited amount of research conducted using this type of data as well as the limitations of recording devices in terms of the granularity signal measurement. 
Since consumer-grade wearable devices for health tracking increasingly allow the capture of real-time physiological signals (e.g Empatica E4 wristband, Fitbit sensors, Garmin watches), researchers are now able to gather multi-sensor-source datasets as input for the study of developing automatic emotion recognition system and automatic stress detection models. 
Despite the possibilities, there has been a limited amount of research in this field to date due to three commonly-known challenging mentioned in \cite{GJORESKI2017159}.
There are two conventional approaches to building stress detection models, which correspond to two different training methods: subject-independent models and subject-dependent models.   
The hypothesis of the subject-dependent stress detection model is that the physiological response to stress stimuli is different for each individual and the stress monitoring systems need to adapt to the stress pattern of each person \cite{Schmidt2018WearableAA}. 
Therefore, stress detection models are likely to perform more accurately when they are trained with each individual's data instead of using external data from various individuals.
Nkurikiyeyezu et al. found that this hypothesis holds true by comparing the accuracy score of both subject-dependent and subject-independent stress detection models trained on high-resolution EDA and ECG signals using the SWELL \cite{SWELLKoldijk2014} and WESAD \cite{Schmidt2018IntroducingWA} datasets \cite{Nkurikiyeyezu2019TheIO}.
Hence, most research to date has concentrated on the application of subject-dependent stress detection models while ignoring subject-independent models despite their practicality and cost-efficiency for consumer-grade application scenarios.
In order to address these issues, we investigate the usage of physiological data recorded from consumer-grade wearable devices for automatic stress detection and propose a new model that improves the accuracy of subject-independent stress detection. 
In summary, we present three main contributions of this paper:
\begin{enumerate}
    \item Through extensive experiments, we prove that the fusion of multiple sensor sources of the consumer-grade wearable device can enhance the accuracy of stress detection models compared to when using each signal individually. 
    \item We propose a bio-signal processing pipeline with a novel training method for the subject-independent models that learn stress/non-stress patterns of EDA, BVP, ST, and their fusion. 
    \item Our proposed model outperforms traditional Machine Learning methods as well as being $1.63\%$ more accurate than the state-of-the-art model on the same experiment dataset. 
\end{enumerate}

\section{Related Work}
Various human contextual data sources, such as Heart rate (HR), Heart Rate Variability (HRV), and Electrodermal Activity (EDA) are found to be discriminative signals for stress level measurement \cite{Can2019ContinuousSD}. 
Using such multi-modal contextual signals,  Nkurikiyeyezu et al.,  Schmidt et al., and  Siirtola manage to build high-accuracy subject-dependent stress detection models \cite{Siirtola2020ComparisonOR,Nkurikiyeyezu2019TheIO,Schmidt2018IntroducingWA}.
Results of these works show that subject-dependent models outperformed subject-independent ones and the gap between the performance of the two models is huge.

In 2018, Schmidt et al. released a public multimodal dataset named WESAD which captured both high-resolution and low-resolution physiological contextual signals of 15 participants under different conditions  \cite{Schmidt2018IntroducingWA}.
They also provided preliminary work on their dataset by training a subject-independent stress detection model.
In a binary classification task, they achieved an average accuracy score of $88.33\%$ ($0.25$) using Random Forest classifiers trained on combinations of low-resolution sensor signals (EDA, BVP, TEMP).
However, as the number of stress and non-stress samples in the WESAD dataset is unequal, this accuracy score cannot reflect the stress detection capability of the model completely as the model can achieve a high accuracy score by predicting the value of the majority class for all predictions.

Siirtola continued to evaluate the performance of subject-independent stress detection models using the same features as in the preliminary work of Schmidt et al. but with another appropriate evaluation metric and different window size \cite{Siirtola2019ContinuousSD}.
The best model using Linear Discriminant Analysis (LDA) trained on three signals which include Skin Temperature (ST), BVP, and HR achieves the highest average balanced accuracy score of $87.4\%$ ($10.4$).
This result is high for subject-independent stress detection model.
However, they suggested that the significant variation in recognition accuracy between study subjects can be alleviated by building subject-dependent stress detection instead.

In 2019, Nkurikiyeyezu et al. compared the performance of these two models using high-resolution EDA and HRV signals from both WESAD \cite{Schmidt2018IntroducingWA} and SWELL \cite{SWELLKoldijk2014} dataset \cite{Nkurikiyeyezu2019TheIO}.
The accuracy score was chosen as the appropriate evaluation metric for their works since they down-sampled the dataset randomly to balance the number of samples in both classes. 
They provided evidence that the subject-dependent stress detection model outperformed the subject-independent one.
They also proposed a hybrid calibrated model to improve the performance of the subject-independent model from $42.5\% \pm 19.9\%$ to $95.2\% \pm 0.5\%$ by including a small number of samples of the unseen subject $(n = 100)$ \cite{Nkurikiyeyezu2019TheIO}.
However, their proposed hybrid calibrated model is not an enhanced version of the subject-independent model but a different way for subject-dependent model training as it requires a small amount of stress/non-stress annotated samples from the targeted subject. 
Additionally, their work was only limited to the use of high-resolution signals, which are usually recorded using laboratory devices only, without analysing the performance of their models on low-resolution signals captured from consumer-grade wearable devices. 
More work is needed to analyse the possibility of improving the performance of the subject-independent stress detection model trained on low-resolution physiological signals. 

\section{Stress Detection Dataset}
To improve the stress prediction accuracy of the subject-independent model, we conducted experiments on the benchmark dataset that is used extensively in related works \cite{Nkurikiyeyezu2019TheIO,Siirtola2019ContinuousSD,10.1145/3460418.3479320} to evaluate the performance of stress detection model. 
The benchmark dataset named WESAD \cite{Schmidt2018IntroducingWA} consists of four different types of low-resolution physiological data collected from 15 participants under two different study protocols in a laboratory environment.
The low-resolution physiological signals including accelerometer (ACC), skin temperature (ST), Blood Volume Pulse (BVP), and Electrodermal Activity (EDA) are recorded using the Empatica E4 medical-grade wearable sensor, which facilitates real-time physiological data acquisition regardless of user context.
Among the four signals, only three bio-signals which are related directly to the response of acute stress includes EDA, BVP, and ST.
In our work, we concentrate on analysing the use of low-resolution EDA, BVP, and ST signals, which are recorded with a sampling rate of 4 Hz, 64 Hz, and 4 Hz respectively, to improve the stress prediction accuracy of a subject-independent model.
Each study protocol in the dataset comprises of amusement, stress, meditation, and baseline conditions in different orders for each participant. 
Details of the four affective conditions are as follows:
\begin{enumerate}
    \item \textbf{Baseline Condition}: This condition lasts for 20 minutes which aims to capture the neutral affective state of the participant. The participant is asked to sit or stand at a table with neutral reading material. 
    \item \textbf{Amusement Condition}: The participant watches a set of eleven funny video clips. A short neutral time period of 5 seconds is presented between the video clips. The total length for this condition is 392 seconds.
    \item \textbf{Stress Condition}: The participant is exposed to the Trier Social Stress Test (TSST) \cite{Kirschbaum1993TheS}, where they are required to provide a five-minute speech on their strengths and weaknesses in front of a panel that  includes three human resource specialists. Finally, the participant counts down from 2023 in decrements of 17, and is requested to start over if they make a mistake. The total length of this condition is about 10 minutes.
\end{enumerate}

The total duration of the study protocol is about two hours, which is considered to be long enough to capture sufficient physiological data for stress detection model training.
Since previous works using this dataset employ study protocol is used as the ground-truth for constructing the stress detection model
\cite{Nkurikiyeyezu2019TheIO,Schmidt2018IntroducingWA,Siirtola2019ContinuousSD,Ninh2021AnalysingStress}, we also use the same ground-truth as in previous works for consistent comparison of stress prediction accuracy of the proposed models. 
In detail, the baseline and amusement condition are classified into non-stress class while the stress condition is considered as the stress one. 

\section{Experiments Description}
In this research, we employ the bio-signals of the WESAD dataset, EDA, BVP, and ST signals, that can be recorded from separate sensors integrated on a low-cost consumer-grade wearable device to predict the stress pattern of the individual. 
We also propose a bio-signal processing pipeline for each signal individually before extracting statistical features, which is described in \ref{sst:FeatureExtraction}. 
Several statistical features identified in other researcher's findings are extracted from these signals, and then concatenated together to build our prediction models. 
We then conduct many experiments with different training approaches to evaluate the effectiveness of combining signals in subject-independent models.

\subsection{Bio-signal Processing and Statistical Feature Extraction of EDA, BVP, and ST} \label{sst:FeatureExtraction}
For both EDA and BVP, we extract statistical features using NeuroKit2 package\footnote{https://github.com/neuropsychology/NeuroKit} \cite{Makowski2021neurokit} and HRV-analysis library\footnote{https://github.com/Aura-healthcare/hrv-analysis} for each 60-second segment. 
The window shift used in our experiment is 0.25 second.
The values of the window size and window shift are the same as in the original paper of WESAD dataset for consistency when comparing the prediction results of the models \cite{Schmidt2018IntroducingWA}.
As the physiological signals vary from person to person, we employ feature normalisation method to reduce the difference people's physiological responses.
In addition, since the signals recorded using consumer-grade wearable device such as EDA, BVP, etc. contain many types of noise, we utilise many signal processing techniques to remove noises, baseline drifts, and outliers in the raw signal.
These steps are combined together to clean the raw signal before extracting statistical feature, which is considered to be a bio-signal processing flow to improve the quality of the extracted feature.

For the EDA, the raw signal in each 60-second segment is firstly pre-processed to remove motion artifacts using the wavelet-based adaptive denoising procedure as described in \cite{Chen2015WaveletbasedMA}.
The signal is then filtered by a fourth-order Butterworth low-pass filter with cut off frequency of 0.5 Hz to remove line noise. The min-max normalization is then applied to the cleaned signal to remove the inter-individual difference before it is inputted into the NeuroKit2 package to decompose into Skin Conductance Response (SCR) and Skin Conductance Level (SCL) using the cvxEDA method \cite{7229284}.
Other characteristics of SCR including SCR Peaks, SCR Onsets, and SCR Amplitude are also extracted.
Finally, the statistical EDA features from three related works \cite{Choi2012DevelopmentAE,Nkurikiyeyezu2019TheIO,Schmidt2018IntroducingWA} are computed which resulted in a 36-dimensional vector.

For the BVP, we firstly clean the raw signal in each window segment by removing the outlier values over the $98^{\text{th}}$ and below the $2^{\text{th}}$ percentile using winsorisation method as in \cite{phdthesisMartin} and removing the baseline drift using Butterworth high-pass filter with cut-off frequency of $0.5$ Hz as in \cite{Kher2019SignalPT}. We then apply min-max normalization to the cleaned signal to minimise the physiological signal difference between individuals before following the previous research \cite{Ninh2021AnalysingStress} to employ the Elgandi processing pipleline \cite{Elgendi2013SystolicPD} to clean the photoplethysmogram (PPG) signal \cite{Nabian2018AnOF} and detect systolic peaks.
The systolic peaks are used to compute a list of RR-intervals, which are then pre-processed using the hrv-analysis package to remove outliers and ectopic beats \cite{Kamath1995Correction} as well as interpolating missing values.
The cleaned RR-intervals are used to compute the NN-intervals, which are main items to compute time-domain, frequency-domain, geometrical, and Poincare-plot features.
For frequency-domain HRV features, we employ the same parameters of low (LF: 0.04-0.15 Hz) and high (HF: 0.15-0.4 Hz) frequency bands as in \cite{Schmidt2018IntroducingWA}.
The range of very-low frequency band used in our work is the same as in HRV-analysis package (0.003-0.04 Hz).  
In summary, we inherit most of the HRV features from \cite{Nkurikiyeyezu2019TheIO,Schmidt2018IntroducingWA} and combined them into a 30-dimensional vector. 

For the ST, the statistical features are extracted on the raw 60-second segment signal as in \cite{Schmidt2018IntroducingWA}.
The fusion of statistical features from three signal-sources is a 72-dimensional vector.
The detail of extracted features is shown in Table \ref{tab:ExtractedFeatures}.

\begin{table}[ht!]
    \centering
    \caption{List of extracted features. Abbreviations: \# = number of, $\sum$ = sum of, STD = standard deviation, RMS = Root Mean Square.}
    \begin{tabular}[c]{| c | p{5.5cm} | p{5.5cm} |} 
     \hline
      & \textbf{Feature} & \textbf{Description}\\
     \hline
     \multirow{22}{3em}{EDA} & 
        $\mu_{EDA}$, $\sigma_{EDA}$, $\min_{EDA}$, $\max_{EDA}$ & Mean, STD, min, max of the EDA\\
        & $\partial_{EDA}$ & Slope of the EDA\\ 
        & $\text{range}_{EDA}$, $\text{range}_{SCR}$ & Dynamic range of EDA and SCR\\
        & $\mu_{SCL}$, $\sigma_{SCL}$ & Mean, STD of the SCL \\
        & $\text{corr}(SCL, t)$ & Correlation btw SCL and time \\
        & $\#_{Peak}$ & $\#$ identified SCR peaks \\
        & $\sum_{SCR}^{Amp}$, $\sum_{SCR}^{t}$ & $\sum$ SCR startle magnitudes and response durations \\
        & $\int_{SCR}$ & Area under the identified SCRs \\
        & $\mu_{SCR}$, $\sigma_{SCR}$, $\max_{SCR}$, $\min_{SCR}$ & Mean, STD, min, max of the SCR \\
        & $\mu_{\nabla_{SCR}}$, $\sigma_{\nabla_{SCR}}$, $\mu_{\nabla(\nabla_{SCR})}$, $\sigma_{\nabla(\nabla_{SCR})}$ & Mean and STD of the 1st and second derivative of the SCR \\
        & $\mu_{Peak}$, $\sigma_{Peak}$, $\max_{Peak}$, $\min_{Peak}$ & Mean, STD, min, max of SCR Peaks \\
        & $\text{kurtosis}(SCR)$, $\text{skewness}(SCR)$ & Kurtosis and skewness of SCR\\
        & $\mu_{Onset}$, $\sigma_{Onset}$, $\max_{Onset}$, $\min_{Onset}$ & Mean, STD, min, max of SCR Onsets \\
        & ALSC = $\sum_{n=2}^{N}\sqrt{1 + (r[n] - r[n-1])^2}$ & Arc length of the SCR \\
        & INSC = $\sum_{n=1}^{N}\left|r[n]\right|$ & Integral of the SCR \\
        & APSC = $\dfrac{1}{N}\sum_{n=1}^{N}r[n]^2$ & Normalized average power of the SCR \\
        & RMSC = $\sqrt{\dfrac{1}{N}\sum_{n=1}^{N}r[n]^2}$ & Normalized RMS of the SCR \\
     \hline
     \multirow{21}{3em}{BVP} 
     & $\mu_{HR}$, $\sigma_{HR}$ & Mean and STD of Heart Rate \\
     & $\mu_{HRV}$, $\sigma_{HRV}$ & Mean and STD of HRV \\
     & $\text{kurtosis}(HRV)$, $\text{skewness}(HRV)$ & Kurtosis and Skewness of HRV \\
     & $f^{VLF}_{HRV}$, $f^{LF}_{HRV}$, $f^{HF}_{HRV}$, $f^{LFNorm}_{HRV}$, $f^{HFNorm}_{HRV}$ & Very low (VLF), Low (LF), and High (HF) frequency band in the HRV power spectrum and their normalized values. \\
     & $f^{LF/HF}_{HRV}$ &  Ratio of HRV LF and HRV HF. \\
     & $\sum\limits_{x \in \{\text{VLF, LF, HF}\}}^f$ &  $\sum$ of the freq. components in VLF-HF\\
     & $\text{NN}50$, $\text{pNN}50$, $\text{NN}20$ $\text{pNN}20$ & \# and percentage of HRV intervals differing more than 50 ms and 20ms. \\
     & HTI & HRV Triangular index \\
     & $\text{rms}_{HRV}$ & RMS of the HRV\\
     & SD1, SD2 & Short and long-term poincare plot descriptor of HRV \\
     & RMSSD, SDSD & RMS and STD of all interval of differences between adjacent RR intervals.\\
     & SDSD\_RMSSD & Ratio of SDSD over RMSSD.\\
     & RELATIVE\_RR ($\mu$, median, $\sigma$, RMSSD, kurtosis, skewness) & Mean, median, STD, RMSSD, kurtosis, and skewness of the relative RR. \\ 
    \hline
    \multirow{2}{3em}{ST} & $\mu_{ST}$, $\sigma_{ST}$, $\min_{ST}$, $\max_{ST}$ & Mean, STD, min, max of ST \\
    & $\text{range}_{ST}$, $\partial_{ST}$ & Range and slope of ST \\
    \hline
    \end{tabular}
    \label{tab:ExtractedFeatures}
\end{table}

\subsection{Stress Detection Model Training Methodology}
\begin{figure}[t!]
\centering
\includegraphics[width=1.0\linewidth]{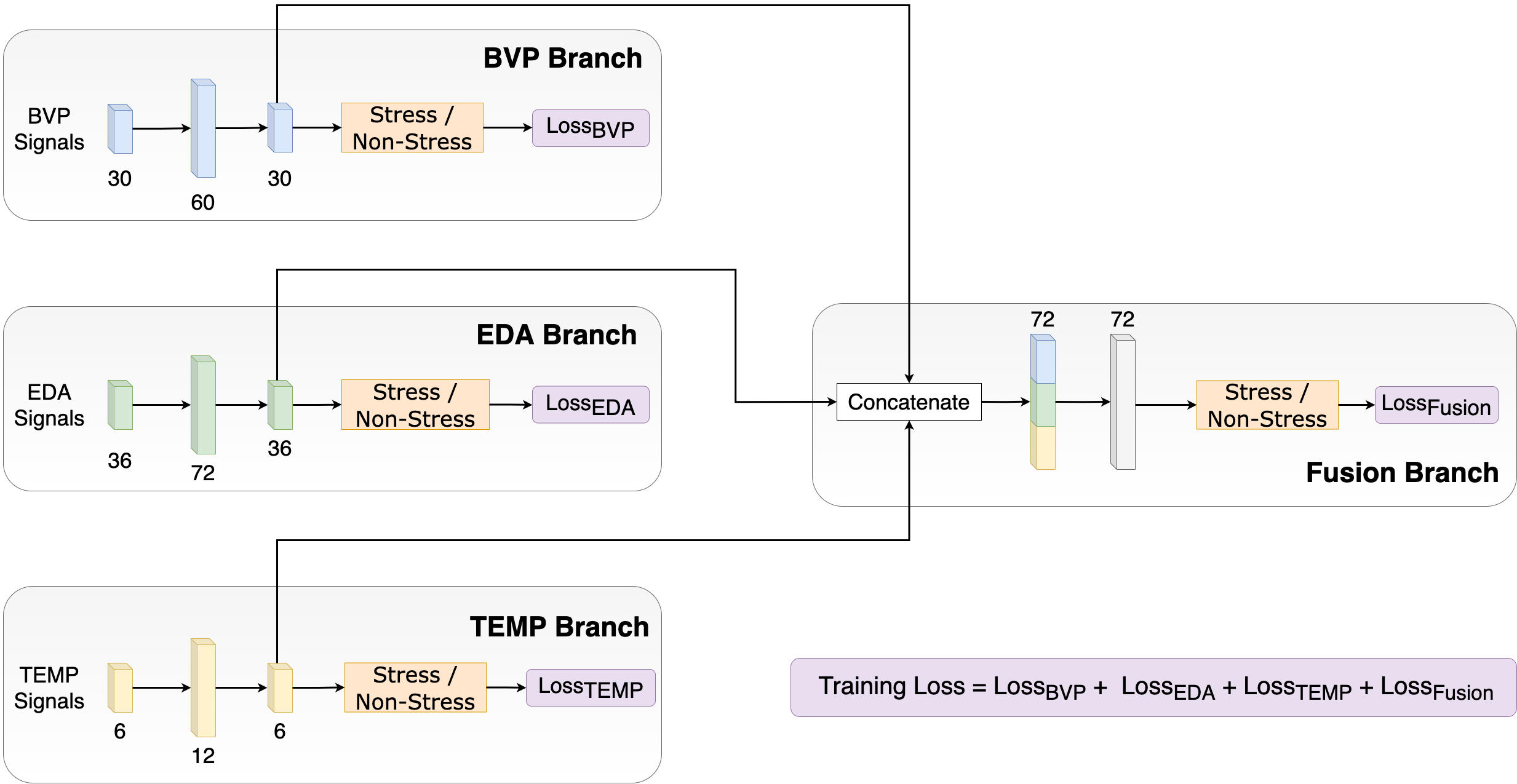}
\caption{The structure of our proposed neural network model. The numbers indicate the number of features of a signal and also the number of units in each layer.}
\label{fig:nnmodel}
\end{figure}

In this research, we build different classifiers to detect the stress condition of each participant in the WESAD dataset. 
Two conventional machine learning classifiers that are widely applied in this field which are Random Forest (RF) and Support Vector Machine (SVM) are applied as baseline models using our proposed feature extraction pipeline. The feature used for these machine learning models is either a feature vector combined from three signals (dimension of 72) or a feature vector of each signal only (dimension of 30 for BVP, 36 for EDA, and 6 for ST).   
Additionally, we introduce a neural network (NN) architecture that captures not only the local detail of EDA, BVP, and ST separately but also the fusion of these signals. 
The neural network model, as depicted in Figure \ref{fig:nnmodel}, contains three distinct embedding modules for each signal then a concatenating layer to learn the joint encoded features. 
The model is then added with three different classification layers for three branches which aims to optimise the performance of embedding stages of EDA, BVP, and ST signals prior to the concatenating step. 
The overall loss used to train the NN model is the sum of losses of all branches. 
In the testing phase, the NN model makes a prediction based on the average of all branches in order to gather the detail of each signal and their combined information. 
We also integrate batch normalization and dropout techniques to make the model converge faster and to address any over-fitting concerns.

We train two models (RF and SVM) using the Leave-One-Subject-Out (LOSO) scheme as in \cite{Schmidt2018IntroducingWA}.
For the NN model, the Leave-One-Subject-Out (LOSO) scheme is also employed, however, the data is split into train set ($80\%$) and validation set ($20\%$) using the Stratified Shuffle Split to deal with the imbalanced nature of the ground-truth distribution in the dataset. 
Additionally, the effectiveness of combining signals in stress detection is also considered. 
For each scenario described above, we run four trials either using features of each signal individually or using the fusion features from multiple sources. 
In the first three trials, the NN model is trained and tested only with its respective branch as illustrated in Figure \ref{fig:nnmodel}. 
By conducting these runs, we assess the effectiveness of using each signal in the stress prediction problem and whether fusing signals can produce better prediction results.

\subsection{Experimental Configuration}
In our experiments, we set $250$ trees for the RF model with enabled out-of-bag, bootstrap samples, max depth of 8, min sample splits of 2, and min sample leaves of 4.
The radial basis function kernel was used in the SVM model with regularization parameter of 10.
The remaining parameters of both RF and SVM model are kept default as in sklearn library\footnote{https://scikit-learn.org/stable/}. 
Both models are set up to take the imbalance of the dataset into account by enabling balanced weights for each class when training. 
These are configurations that achieve the highest accuracy score after we conduct extensive experiments.
The NN model is trained with an Adam optimiser \cite{DBLP:journals/corr/KingmaB14} with a learning rate of $0.003$ while the dropout level and the batch size are set at $10\%$ and 2048 accordingly.  
Regarding the evaluation metrics, we report both balanced accuracy and accuracy scores due to the imbalance between number of samples in the two classes in the dataset. 
Based on the analysis of Straube et al., balanced accuracy (BA) is both an appropriate choice and an intuitive metric to evaluate prediction results of a binary classification problem when an imbalanced dataset is used for testing \cite{Straube2014EvaluationMetric}.
The balanced accuracy is calculated by taking the average of the recall measured on each class separately.

\section{Results}


In Table \ref{table:combineftACC} and Table \ref{table:combineftBA}, we show the effect of combining statistical features from three sources of signal and compare the results of our proposed stress detection model with previous works. 
As can be seen from both tables, all models of ours obtain higher evaluation scores when using the combination features from multiple sources of signals than using only each of them individually.
According to Table \ref{table:combineftACC}, compared to the original work in \cite{Schmidt2018IntroducingWA}, employing our proposed bio-signal processing pipeline before feature extraction step increases the mean accuracy score of conventional Machine Learning model (RF model) around $3.2\%$.
Our proposed NN model improves the performance of state-of-the-art (SOTA) subject-independent stress detection model around $1.63\%$ using the same number of bio-signals.
However, conventional Machine Learning models also achieve competitive accuracy scores compared to the SOTA model when using the fusion feature of three sensor sources with appropriate signal processing before feature extraction.
The mean accuracy scores of our SVM and RF models using the combined features as input are $92.71\%$ and $91.53\%$ with standard deviation of $\pm 7.90\%$ and $\pm 7.24\%$ respectively while the one of our NN model is $94.50\%$ $(\pm 5.64\%)$.
These results indicate that our proposed NN model not only increases the prediction accuracy of subject-independent stress detection model in average, but it also does not result in a large difference of accuracy score between each subject's model.  

\begingroup
\setlength{\tabcolsep}{5.8pt} 
\renewcommand{\arraystretch}{1.05} 
\begin{table}[t!]
\centering
\caption{Comparison of the mean accuracy score between different subject-independent stress detection models in previous works and ours using biosensor signals (window size = 60s, window shift = 0.25s).}
\begin{tabular}{|c|c|c|c|c|c|c|} 
\hline
\multirow{2}{*}{\textbf{Sensor Combinations}} & \multicolumn{2}{c|}{\textbf{Schmidt \cite{Schmidt2018IntroducingWA}}} & \multicolumn{1}{c|}{\textbf{Lam \cite{10.1145/3460418.3479320}}} & \multicolumn{3}{c|}{\textbf{Proposed}} \\
\cline{2-7}
& RF & LDA & StressNAS & SVM & RF & NN \\
\hline
EDA+BVP+ST & 88.33 & 86.46 & 92.87 & 92.71 & 91.53 & \textbf{94.50}\\
EDA & 76.29 & 78.08 & \textbf{79.24} & 76.32 & 75.53 & 77.57 \\ 
BVP & 84.18 & 85.83 & 81.16 & 86.95 & 87.39 & \textbf{89.94}  \\ 
ST & 67.82 & 69.24 & 71.46 & 69.21 & 69.23 & \textbf{74.07}\\ 
\hline
\end{tabular}
\label{table:combineftACC}
\end{table}
\endgroup

In terms of imbalanced-data insensitive evaluation metrics, we report the balanced accuracy scores of our models and compare them with corresponding related work \cite{Siirtola2019ContinuousSD}.
For consistency in comparison the balanced accuracy score with \cite{Siirtola2019ContinuousSD}, we use the window size and window shift of 120 seconds and 0.25 second respectively.
According to the results in Table \ref{table:combineftBA}, our proposed NN model outperforms conventional Machine Learning approaches reported in \cite{Siirtola2019ContinuousSD}.
In detail, the balanced accuracy score of our NN models is $94.16\%$ $(\pm 6.90\%)$, which is higher than the QDA model of \cite{Siirtola2019ContinuousSD} around $12.56\%$ using the same number of bio-signals.
In addition, our RF model achieves higher balanced score than the one in \cite{Siirtola2019ContinuousSD} approximately $12.09\%$, which proves that our bio-signal processing pipeline is efficient and necessary for feature extraction step to enhance subject-independent stress detection model.
The balanced accuracy score of our SVM and RF model trained with features combined from different sensor sources are $93.36\%$ $(\pm 9.14\%)$ and $93.03$ $(\pm 10.19\%)$ respectively.

To facilitate for other future research to compare results with ours, we also report both balanced accuracy and accuracy scores of our models with different settings of window size and window shift that are not reported in Table \ref{table:combineftACC} and Table \ref{table:combineftBA}.
For window size of 120s, the accuracy scores of our SVM, RF, and NN models using combined feature are $95.23$ $(\pm 5.32)$, $94.57$ $(\pm 6.42)$, and $95.26$ $(\pm 4.68)$ correspondingly
For window size of 60s and window shift of 0.25s, the balanced accuracy scores of our SVM, RF, and NN models are $90.85$ $(\pm 9.99)$, $90.32$ $(\pm 11.21)$, and $92.66$ $(\pm 9.06)$ correspondingly. 

\begingroup
\setlength{\tabcolsep}{5.8pt} 
\renewcommand{\arraystretch}{1.05} 
\begin{table}[t!]
\centering
\caption{Comparison of the mean balanced accuracy score between different subject-independent stress detection models in previous work and ours using biosensor signals (window size = 120s, window shift = 0.25s).}
\begin{tabular}{|c|c|c|c|c|c|c|} 
\hline
\multirow{2}{*}{\textbf{Sensor Combinations}} & \multicolumn{3}{c|}{\textbf{Siirtola \cite{Siirtola2019ContinuousSD}}} & \multicolumn{3}{c|}{\textbf{Proposed}} \\
\cline{2-7}
& RF & LDA & QDA & SVM & RF & NN \\
\hline
EDA+BVP+ST & 81.00 & 78.80 & 81.60 & 93.36 & 93.09 & \textbf{94.16}\\
EDA & \textbf{78.30} & 73.50 & 69.70 & 71.34 & 70.12 & 77.52 \\ 
BVP & 81.40 & 81.20 & 67.90 & 87.57 & \textbf{90.92} & 88.28  \\ 
ST & 66.90 & 75.20 & 68.30 & 71.07 & 71.13 & \textbf{77.97}\\ 
\hline
\end{tabular}
\label{table:combineftBA}
\end{table}
\endgroup

\section{Conclusion}
In this paper, we build a model that uses a neural network (NN) architecture for subject-independent stress detection using three types of bio-signals that can be captured from a consumer-grade low-cost device. 
The model contains four different NN modules where each of the three modules learns the embedding features from each bio-signal individually while the remaining one learns the joint embedded feature of the three modules by concatenating the latent-space representation of each signal.
The proposed model is evaluated against Random Forest and Support Vector Machine models on the WESAD dataset using balanced accuracy and accuracy score. 
Our experiments show that using statistical features from multiple sensor sources can produce more accurate stress prediction results.
Additionally, our experiments also show that our proposed NN model outperforms conventional Machine Learning approaches for subject-independent model training for both evaluation metrics.
In detail, our NN model achieves higher evaluation scores than the SOTA model and conventional Machine Learning models while maintaining a low standard deviation score.
We believe that our findings could help promote and guide future efforts in improving subject-independent stress detection models, in order to enhance the integration of stress detection and management system into consumer-grade low-cost wearable devices in a practical and cost-efficient manner.

\subsubsection*{Acknowledgments.} This publication is funded as part of Dublin City University's Research Committee and research grants from Science Foundation Ireland under grant numbers  SFI/13/RC/2106, SFI/13/RC/2106\_P2, and 18/CRT/6223.

\bibliographystyle{splncs04}
\bibliography{bookchapter}

\begin{thebibliography}{10}
\providecommand{\url}[1]{\texttt{#1}}
\providecommand{\urlprefix}{URL }
\providecommand{\doi}[1]{https://doi.org/#1}

\bibitem{Can2019ContinuousSD}
Can, Y.S., Chalabianloo, N., Ekiz, D., Ersoy, C.: Continuous stress detection
  using wearable sensors in real life: Algorithmic programming contest case
  study. Sensors (Basel, Switzerland)  \textbf{19} (2019)

\bibitem{Chen2015WaveletbasedMA}
Chen, W.V., Jaques, N., Taylor, S., Sano, A., Fedor, S., Picard, R.W.:
  Wavelet-based motion artifact removal for electrodermal activity. 2015 37th
  Annual International Conference of the IEEE Engineering in Medicine and
  Biology Society (EMBC) pp. 6223--6226 (2015)

\bibitem{Choi2012DevelopmentAE}
Choi, J., Ahmed, B., Gutierrez-Osuna, R.: Development and evaluation of an
  ambulatory stress monitor based on wearable sensors. IEEE transactions on
  information technology in biomedicine  \textbf{16}(2),  279--286 (2011)

\bibitem{Elgendi2013SystolicPD}
Elgendi, M., Norton, I., Brearley, M., Abbott, D., Schuurmans, D.: Systolic
  peak detection in acceleration photoplethysmograms measured from emergency
  responders in tropical conditions. PLoS One  \textbf{8}(10),  e76585 (2013)

\bibitem{phdthesisMartin}
Gjoreski, M.: Continuos Stress Monitoring using a Wrist Device and a
  Smartphone. Ph.D. thesis, Jožef Stefan International Postgraduate School,
  Ljubljana, Slovenia (09 2016)

\bibitem{GJORESKI2017159}
Gjoreski, M., Luštrek, M., Gams, M., Gjoreski, H.: Monitoring stress with a
  wrist device using context. Journal of Biomedical Informatics  \textbf{73},
  159--170 (2017). \doi{https://doi.org/10.1016/j.jbi.2017.08.006},
  \url{https://www.sciencedirect.com/science/article/pii/S1532046417301855}

\bibitem{7229284}
Greco, A., Valenza, G., Lanata, A., Scilingo, E.P., Citi, L.: cvxeda: A convex
  optimization approach to electrodermal activity processing. IEEE Transactions
  on Biomedical Engineering  \textbf{63}(4),  797--804 (2016).
  \doi{10.1109/TBME.2015.2474131}

\bibitem{Gurrin2014APB}
Gurrin, C., Albatal, R., Joho, H., Ishii, K.: A privacy by design approach to
  lifelogging. In: O'Hara, K., Nguyen, C. and Haynes, P., (eds.) Digital
  Enlightenment Yearbook 2014. pp. 49--73. IOS Press, The Netherlands (2014)

\bibitem{Gurrin2016OverviewON}
Gurrin, C., Joho, H., Hopfgartner, F., Zhou, L., Albatal, R.: Overview of
  ntcir-12 lifelog task. In: NTCIR (2016)

\bibitem{Gurrin2019OverviewOT}
Gurrin, C., Joho, H., Hopfgartner, F., Zhou, L., Ninh, V.T., Le, T.K., Albatal,
  R., Dang-Nguyen, D.T., Healy, G.: Overview of the ntcir-14 lifelog-3 task.
  In: Proceedings of the 14th NTCIR Conference on Evaluation of Information
  Access Technologies (2019)

\bibitem{Gurrin2014LifeloggingPB}
Gurrin, C., Smeaton, A., Doherty, A.: Lifelogging: Personal big data. Found.
  Trends Inf. Retr.  \textbf{8},  1--125 (2014)

\bibitem{Gurrin2021}
Gurrin, C., Joho, H., Hopfgartner, F., Zhou, L., Albatal, R., Healy, G.,
  Nguyen, D.T.D.: Experiments in Lifelog Organisation and Retrieval at NTCIR,
  pp. 187--203. Springer Singapore, Singapore (2021)

\bibitem{10.1145/3460418.3479320}
Huynh, L., Nguyen, T., Nguyen, T., Pirttikangas, S., Siirtola, P.: StressNAS:
  Affect State and Stress Detection Using Neural Architecture Search, p.
  121–125. Association for Computing Machinery, New York, NY, USA (2021),
  \url{https://doi.org/10.1145/3460418.3479320}

\bibitem{Kher2019SignalPT}
Kher, R.: Signal processing techniques for removing noise from ecg signals. In:
  Journal of Biomedical Engineering and Research (2019)

\bibitem{DBLP:journals/corr/KingmaB14}
Kingma, D.P., Ba, J.: Adam: {A} method for stochastic optimization. In: Bengio,
  Y., LeCun, Y. (eds.) 3rd International Conference on Learning
  Representations, {ICLR} 2015, San Diego, CA, USA, May 7-9, 2015, Conference
  Track Proceedings (2015), \url{http://arxiv.org/abs/1412.6980}

\bibitem{Kirschbaum1993TheS}
Kirschbaum, C., Pirke, K.M., Hellhammer, D.H.: The ‘trier social stress
  test’--a tool for investigating psychobiological stress responses in a
  laboratory setting. Neuropsychobiology  \textbf{28}(1-2),  76--81 (1993)

\bibitem{SWELLKoldijk2014}
Koldijk, S., Sappelli, M., Verberne, S., Neerincx, M.A., Kraaij, W.: The swell
  knowledge work dataset for stress and user modeling research. In: Proceedings
  of the 16th International Conference on Multimodal Interaction. p. 291–298.
  ICMI '14, Association for Computing Machinery, New York, NY, USA (2014).
  \doi{10.1145/2663204.2663257}

\bibitem{Makowski2021neurokit}
Makowski, D., Pham, T., Lau, Z.J., Brammer, J.C., Lespinasse, F., Pham, H.,
  Sch{\"o}lzel, C., Chen, S.A.: Neurokit2: A python toolbox for
  neurophysiological signal processing. Behavior Research Methods pp.~1--8
  (2021)

\bibitem{Nabian2018AnOF}
Nabian, M., Yin, Y., Wormwood, J., Quigley, K.S., Barrett, L.F., Ostadabbas,
  S.: An open-source feature extraction tool for the analysis of peripheral
  physiological data. IEEE journal of translational engineering in health and
  medicine  \textbf{6},  1--11 (2018)

\bibitem{Ninh2020OverviewOI}
Ninh, V.T., Le, T.K., Zhou, L., Piras, L., Riegler, M., Halvorsen, P., Lux, M.,
  Tran, M., Gurrin, C., Dang-Nguyen, D.T.: Overview of imageclef lifelog 2020:
  Lifelog moment retrieval and sport performance lifelog. In: CLEF (2020)

\bibitem{Ninh2021AnalysingStress}
Ninh, V.T., Smyth, S., Tran, M.T., Gurrin, C.: Analysing the performance of
  stressdetection models on consumer-grade wearable devices. In: SoMeT (2021)

\bibitem{Nkurikiyeyezu2019TheIO}
Nkurikiyeyezu, K., Yokokubo, A., Lopez, G.: Effect of person-specific
  biometrics in improving generic stress predictive models. Sensors and
  Materials  \textbf{32},  703--722 (02 2020). \doi{10.18494/SAM.2020.2650}

\bibitem{Schmidt2018IntroducingWA}
Schmidt, P., Reiss, A., Duerichen, R., Marberger, C., Van~Laerhoven, K.:
  Introducing wesad, a multimodal dataset for wearable stress and affect
  detection. In: Proceedings of the 20th ACM international conference on
  multimodal interaction. pp. 400--408 (2018)

\bibitem{Schmidt2018WearableAA}
Schmidt, P., Reiss, A., D{\"u}richen, R., Laerhoven, K.V.: Wearable affect and
  stress recognition: A review. ArXiv  \textbf{abs/1811.08854} (2018)

\bibitem{Siirtola2019ContinuousSD}
Siirtola, P.: Continuous stress detection using the sensors of commercial
  smartwatch. Adjunct Proceedings of the 2019 ACM International Joint
  Conference on Pervasive and Ubiquitous Computing and Proceedings of the 2019
  ACM International Symposium on Wearable Computers  (2019)

\bibitem{Siirtola2020ComparisonOR}
Siirtola, P., R{\"o}ning, J.: Comparison of regression and classification
  models for user-independent and personal stress detection. Sensors (Basel,
  Switzerland)  \textbf{20} (2020)

\bibitem{Straube2014EvaluationMetric}
Straube, S., Krell, M.M.: How to evaluate an agent's behavior to infrequent
  events?—reliable performance estimation insensitive to class distribution.
  Frontiers in Computational Neuroscience  \textbf{8}, ~43 (2014)

\bibitem{Kamath1995Correction}
V., K.M., L, F.E.: Correction of the heart rate variability signal for ectopics
  and missing beats. In: Heart Rate Variability, eds M. Malik, Camm A. J (1995)

\end{thebibliography}

\end{document}